\documentclass{article} 
\usepackage{colm2024_conference}

\usepackage{microtype}
\usepackage[hidelinks]{hyperref}
\usepackage{url}
\usepackage{booktabs}

\usepackage{amsmath}
\usepackage{enumitem}
\usepackage{color,soul} 
\usepackage{multirow}
\usepackage{graphicx}
\usepackage{svg}
\usepackage{wrapfig}

\title{Situated Natural Language Explanations}

\author{Zining Zhu$^{1,2,3}$, Haoming Jiang$^1$, Jingfeng Yang$^1$, Sreyashi Nag$^1$, \\
\textbf{Chao Zhang$^{1,4}$, Jie Huang$^{1,5}$, Yifan Gao$^{1}$, Frank Rudzicz$^{2,3,6}$, Bing Yin$^1$} \\
$^1$ Amazon Search, $^2$ University of Toronto, $^3$ Vector Institute of Artificial Intelligence, \\
$^4$ Georgia Institute of Technology, $^5$ University of Illinois at Urbana-Champaign \\
$^6$ Dalhousie University\\
\texttt{zining@cs.toronto.edu}
}

\colmfinalcopy 

\begin{document}

\title{Situated Natural Language Explanations}

\maketitle

\begin{abstract}
Natural language is among the most accessible tools for explaining decisions to humans, and large pretrained language models (PLMs) have demonstrated impressive abilities to generate coherent natural language explanations (NLE). The existing NLE research perspectives do not take the audience into account. An NLE can have high textual quality, but it might not accommodate audiences' needs and preference. To address this limitation, we propose an alternative perspective, \textit{situated} NLE. On the evaluation side, we set up automated evaluation scores. These scores describe the properties of NLEs in lexical, semantic, and pragmatic categories. On the generation side, we identify three prompt engineering techniques and assess their applicability on the situations. Situated NLE provides a perspective and facilitates further research on the generation and evaluation of explanations.\\
\textbf{Warning:} The explanations generated by LLMs may contain factually incorrect statements, and they do not reflect the attitudes of the authors.
\end{abstract}

\section{Introduction}
Increasingly powerful DNN-based decision-masking systems are being deployed to an increasing number of scenarios. However, these systems can be complex and, as a result, there is a need to increase the trustworthiness of the decisions they make. This has led to the development of various methods for explaining the decisions made by these systems \citep{danilevsky-etal-2020-survey,zhang_explainable_2020}. Natural language explanations (NLEs) offer an alternative to traditional explanations, such as attribution scores and rationales, by providing human-readable information. 
\begin{wrapfigure}{r}{.4\linewidth}
    \includegraphics[width=\linewidth]{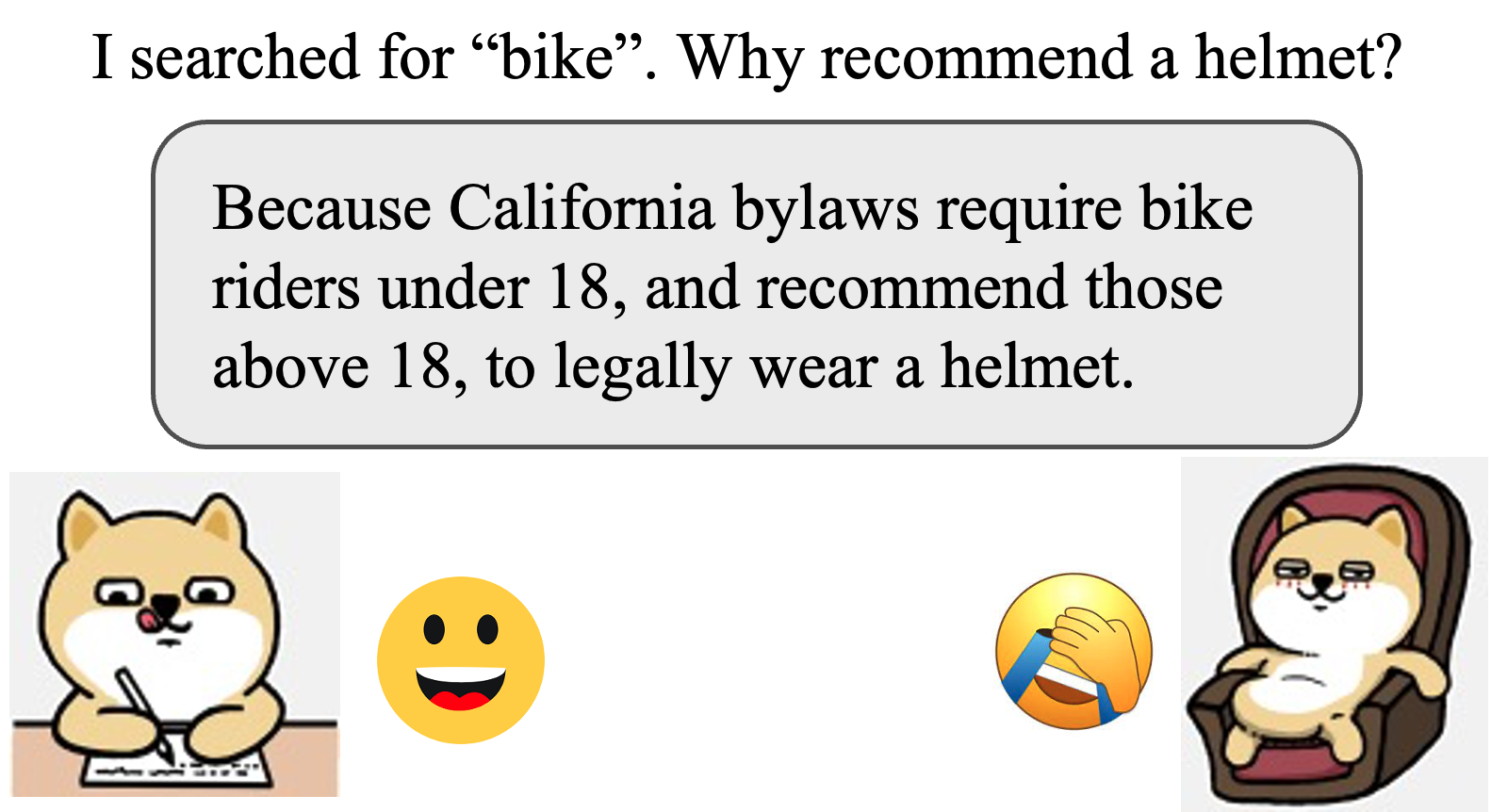}
    \caption{A detailed explanation might be suitable for a careful customer (left) but is not suitable for a customer casually looking around (right) -- we need to adapt the explanations to the situations.}
    \label{fig:nle-motivation}
\end{wrapfigure}
A good NLE should be simple and broad \citep{lombrozo2016Explanatory}. A good NLE should also be accurate, coherent, and informative, and it should help to increase the trustworthiness of the decisions being made \citep{lakkaraju2022rethinking}.

Previous studies have largely focused on the reasoning aspects of NLE. We argue that these static approaches may be inadequate because of the implicit assumption that the quality of an explanation is independent of the audience. However, the quality of an explanation also depends on the audience's needs and preferences \citep{see2019makes,miller2017explanation,achinstein1983nature}. Therefore, in addition to the textual aspects of NLE, we should also incorporate the aspects about the audience when studying the problem of explanation. 

To illustrate the importance of considering the audience when generating explanations, consider the following example. In \textbf{situation 1 (S1)}, a customer is shopping for a bike and is in a hurry, so they only spend a short time searching. In this situation, an explanation such as ``A helmet makes riding bicycles safer'' may be sufficient. In contrast, in \textbf{situation 2 (S2)}, another customer is careful in choosing a product to buy and wants more detailed information to help them make a decision. In this case, the same explanation may not provide enough information to differentiate between different products. A more informative explanation (for example the NLE shown in Figure \ref{fig:nle-motivation}) is more adequate. These examples demonstrate that it is crucial to consider the situation when generating -- and evaluating the quality of -- natural language explanations. 

This paper first aims at establishing an evaluation framework that describes the properties of the NLEs from lexical, semantic, and pragmatic categories. This framework allows us to quantify how the PLMs generate NLEs when prompted in specific techniques.

We also study how to incorporate the audiences' needs and preferences in situations when generating the NLEs using large pre-trained language models (PLMs). These models have demonstrated strong reasoning abilities and are capable of generating high-quality explanations \citep{radford_language_2019,wei2022emergent,mishra_reframing_2022,wiegreffe-etal-2022-reframing}. We propose \textit{situated prompts}, a framework that utilizes either natural language descriptions of the context or desired properties of the explanation to guide the generation of PLMs. We identify three techniques for writing prompts for situated NLEs, and recommend their applications based on the situated evaluation results in an array of settings.

Our contributions can be summarized as follows:
\begin{enumerate}[nosep]
    \item We advocate for the perspective that the generation and evaluation of NLEs should consider the situations of the audience.
    \item A situated evaluation framework with scores in lexical, semantic, and pragmatic categories.
    \item Situated prompts, a convenient prompt engineering and selection method to generate situated NLEs in English using large PLMs.
\end{enumerate}

\section{Related Work}
\paragraph{Natural language explanation} 
Natural languages are especially helpful in explaining problems requiring high-level reasoning. Examples of these problems include inference \citep{camburu_make_2020}, multiple choice questions requiring commonsense \citep{rajani_explain_2019}, question-answering \citep{aggarwal_explanations_2021} and product recommendations \citep{chen2019generate,wen_expscore_2022}. \citet{wiegreffe-marasovic-2021-review} collected a review of these datasets. They implicitly assume a uniform set of criteria for ``good'' NLEs. 
For example, good NLEs should be simple \citep{lombrozo2007simplicity}, clear and informative \citep{clinciu-etal-2021-study}. The situation of the explanations' readers are not part of the problem setting. 

As a clarifying sidenote, we consider explanation to be different from rationale -- the latter (also beneficial in various ways \citep{carton-etal-2020-evaluating}) is a subset of the input tokens, while the former can include external tokens. \citet{jacovi-goldberg-2021-aligning} discusses this issue further. Explanation is almost identical to free-text rationales \citep{wiegreffe2020measuring,chen-etal-2023-rev}, except that free-text rationales are more frequently used in the context when the correctness is at high stakes.

\paragraph{Controllable generation}
Our study is closely related to controllable generation. Applying controls towards various aspects in generating natural languages can be operationalized by imposing constraints to the decoding procedure, for example via reward functions \citep{qin_cold_2022,inoue-etal-2021-summarize}, information bottleneck \citep{paranjape-etal-2020-information}, additional discriminator \citep{krause-etal-2021-gedi-generative} or Monte-Carlo tree search \citep{ferreira2022controlling}. 
Another controllable generation avenue lets the DNN models respond to external ``control signals''. Examples of models in this category include diffusion models \citep{li2022diffusion} and Transformers \citep{keskar_ctrl_2019}.
We defer to \citet{wang2022controllable} for a more comprehensive review of controllable generation methods. Situated prompts belong to the second category, but instead of training new DNNs, we leverage the existing interfaces (prompts) to control the PLMs to generate natural languages without fine-tuning. 


\paragraph{Human-centered explainable AI}
Researchers have proposed numerous methods to make AI more explainable, and most of them focus on mechanistic approaches. Their utilities are subject to increasing concerns. Multiple researchers called for more consideration of the human factors when explaining AI \citep{liao2022connecting,boyd-graber-etal-2022-human,yeung2020,miller_explanation_2019}. Our methodological framework expands the discussion of human-centered explanations to natural language explanations.

\paragraph{Pragmatics and Theory of Mind}
The research toward human-centered natural language explanations is related to the research of pragmatics. We defer to \citet{fried_pragmatics_2022} for a recent summary of the resources and studies on the pragmatic factors of languages. Recently, several works also found that large PLMs show some capabilities of human-like Theory-of-Mind modeling, but their performances were still far below those of humans \citep{Trott2022know,sap2022neural,qiu2021towards}.

\section{Situated Evaluation}
\label{sec:situated-evaluation}
To quantitatively evaluate the explanations and their suitability to the situations, we propose a \textit{situated evaluation} framework, where several automatic metrics characterize the NLEs from multiple aspects. Automatic evaluation metrics are widely used in the language generation and explanation literature \citep{carton-etal-2020-evaluating}. Here we consider metrics that describe the generated texts along the lexical, semantic, and pragmatic dimensions. To evaluate the NLEs in various situations, we can focus on different metrics -- We briefly describe each metric as follows:

\subsection{Lexical category}
\paragraph{Length} We consider the token length\footnote{tokenized by NLTK's word\_tokenize} of the generated explanations. Note that the length can be controlled by modifying the generation pipeline (e.g., stop the generation procedure once hitting a set of ``termination tokens'').

\paragraph{Self-BLEU} \citet{zhu2018texygen} proposed self-BLEU to measure the self-repetitiveness of the generation. A higher self-BLEU implies a lower diversity: Each explanation has more overlapping with other explanations. In an extreme case, producing a dummy explanation, ``It is what it is'' to all problems, will result in a self-BLEU of $1.0$.

\paragraph{Type-token ratio (TTR)} This score measures the lexical richness of the generation. Type-token ratio is usually used in automatic essay scoring \citep{ramesh2021automated} and text-based diagnostic predictions \citep{zhu2018semi}. The maximum possible TTR is $1.0$, acquired when all words are unique in this explanation. TTR is usually negatively correlated to self-BLEU.

\subsection{Semantic category}
\paragraph{BLEU score} \citet{papineni-etal-2002-bleu} proposed BLEU score, which can be used to evaluate the similarity on the n-gram level between the human-written explanations. BLEU score only applies to the datasets with human-provided explanations.

\paragraph{Concreteness} This is the mean concreteness rating (which can be looked up in a table compiled by \citet{koper-schulte-im-walde-2017-improving}) of all words in the explanation, scaled to between 0 and 1. A larger value indicates more concrete choices of words, and vice versa. We refer to this metric as ``concreteness'' for short.

\subsection{Pragmatic category}
\paragraph{Convincingness} Recent literature shows that large PLMs are capable of producing convincing discourse \citep{lin-etal-2022-truthfulqa}, which means they are able to compute reasonable scores to evaluate the convincingness of texts with suitable test form. The \href{https://github.com/google/BIG-bench/tree/main/bigbench/benchmark_tasks/convinceme}{Convince Me} task at BigBench \citep{srivastava_beyond_2022} discusses additional details. Motivated by the analyses, we set up a cloze test that queries a ``tester'' PLM whether the given explanation is convincing: 

S1: \textit{Q\_A. Is this choice convincing? $\langle$mask$\rangle$.}

S2: \textit{Q\_A because \textit{Explanation}. Is this choice convincing? $\langle$mask$\rangle$.}

S3: \textit{Q\_A because it is what it is. Is this explanation convincing? $\langle$mask$\rangle$.}\footnote{``it is what it is'' is a dummy explanation.}

S4: \textit{Q\_A because \textit{Explanation}. Is this explanation convincing? $\langle$mask$\rangle$.}
    
Here the \textit{Q\_A} is the problem to be explained. The format is decided based on the dataset -- \S \ref{subsec:data} contains more details. For each of S1-S4, we use a powerful PLM, \texttt{roberta-large}\footnote{We also tried several other models including bert-base-cased, xlm-roberta-large, gpt-j-6B on a small set of $30$ problems from \citet{rajani_explain_2019}. The convincingness results of roberta-large correlated to our own annotations the most.} as the ``tester'' that computes the logits of filling in ``yes'' and ``no'' to the $\langle$mask$\rangle$ token. Let $P_i(\textrm{yes})$ and $P_i(\textrm{no})$, $i\in \{1,2,3,4\}$, denote the logits for S$i$. 
The logit values are affected heavily by other tokens in the vocabulary. To focus on the preference of the tester PLM along the ``yes''-``no'' spectrum, we take the softmax of the two logits for each sentence: 

\begin{equation*}
    y_i = \frac{e^{P_i(\textrm{yes})}}{e^{P_i(\textrm{yes})} + e^{P_i(\textrm{no})}}.
\end{equation*}
Then, we compute two scores: 
\begin{equation}
    \textrm{C\_v\_no} = \frac{1}{2}(y_2 - y_1 + 1).
\end{equation}
This is ``convincingness versus no explanation'', answering if the explanation improves the convincingness of the \textit{choice}, compared to no explanation.
\begin{equation}
    \textrm{C\_v\_dummy} = \frac{1}{2}(y_4 - y_3 + 1).
\end{equation}
This is ``convincingness versus dummy'', which answers if the explanation \textit{itself} is more convincing, compared to a dummy explanation. Both convincingness scores range from 0 to 1. A larger value corresponds to a higher likelihood that the belief of a ``tester'' is changed by the explanation, using a ``tester'' PLM as a proxy.
    
\paragraph{Acceptability} \citet{wiegreffe-etal-2022-reframing} collected human ratings of acceptability of explanations towards commonsense QA problems and released model checkpoints. We use their \texttt{t5-large} checkpoint to predict the acceptability of explanations. The acceptability score also ranges from 0 to 1. Intuitively, a higher value indicates that the explanation ``matches the facts'' in the common sense involved in this problem.

\begin{table*}
    \centering
    \resizebox{\linewidth}{!}{
    \begin{tabular}{l l l l}
    \toprule 
        Question and Answer & Model and Situation & Explanation & Scores \\ \midrule 
        \multirow{18}{4cm}{Q: ``There are 10 apples on an 
        apple tree. Three fall off. Now there are X apples.'' \\ What is this an example of? \\
        \vspace{0.5cm} Choices:\\ \hspace{1em}- park\\
        \hspace{1em}- coloring book\\
        \hspace{1em}- garden center\\
        \hspace{1em}- \textul{math problem}\\
        \hspace{1em}- gravity} & \multirow{4}{3cm}{Model: ChatGPT\\ Situation: busy} & \multirow{4}{7cm}{it involves counting and determining the new quantity of apples.} & Concreteness: 0.212 \\
        & & & C\_v\_no: 0.571 \\ 
        & & & C\_v\_dummy: 0.554 \\ 
        & & & Acceptability: 0.187 \\ \cline{2-4}
        & \multirow{4}{3cm}{Model: ChatGPT\\ Situation: detailed} & \multirow{4}{7cm}{this scenario involves a mathematical operation of subtracting 3 from 10 to find the remaining number of apples on the tree.} & Concreteness: 0.270\\
        & & & C\_v\_no: 0.523 \\
        & & & C\_v\_dummy: 0.514 \\
        & & & Acceptability: 0.494 \\ \cline{2-4}
        & \multirow{4}{3cm}{Model: Cohere Command\\ Situation: busy} & \multirow{4}{7cm}{this involves solving for x, an unknown variable, answering a question about quantity which is typically associated with mathematics and math problems.} & Concreteness: 0.185 \\
        & & & C\_v\_no: 0.537 \\
        & & & C\_v\_dummy: 0.482 \\
        & & & Acceptability: 0.110 \\ \cline{2-4}
        & \multirow{5}{3cm}{Model: Cohere Command \\ Situation: detailed} & \multirow{6}{7cm}{This is best described as a math problem because the description includes a quantitative situation that deals with numbers, which is a primary focus in mathematics. If you would like, I can break down the problem for you to better illustrate how it is a mathematical concept.} & Concreteness: 0.230\\
        & & & C\_v\_no: 0.742\\
        & & & C\_v\_dummy: 0.707 \\
        & & & Acceptability: 0.187\\ 
        & & & \\
        & & & \\
        & & & \\
    \bottomrule 
    \end{tabular}
    }
    \caption{Examples of PLM-generated NLEs. All settings are 1F\_2F\_3F (as defined in \S \ref{subsec:prompting-techniques}) for convenience.}
    \label{tab:examples-chat-cohere}
\end{table*}

\section{Situated Explanation in Natural Languages}
\label{sec:method}

\subsection{Prompting PLMs for explanation} 
\label{subsec:method-generation}
PLMs can store a massive amount of external information \citep{burns2022discovering} and can assemble pieces of information into meaningful discourse. These capacities allow them to generate NLEs that can be comparable to, or even better than human-written NLEs in some aspects \citep{wiegreffe-etal-2022-reframing}. 
There are many ways we can prompt a PLM to generate NLEs. One popular approach is specified by \citet{rajani_explain_2019}:

\vspace{0.5em}
\noindent \fbox{
\parbox{\linewidth} {
    Prompt: \textit{Q\_A because} 
    }
}

This prompt appends a term, ``because'' to the end of Q\_A, the question, choices and answer. In general, Q\_A is the \textit{explanandum}, the text describing the phenomenon that we want to explain. What follows ``because'' is the NLE. We refer to this prompt as the ``base prompt'' henceforth.

\subsection{Techniques for writing a situated prompt}
\label{subsec:prompting-techniques}
We consider three techniques for writing prompts for PLMs to generate situated NLEs.

\paragraph{1. Specify the desired feature} This may be the most intuitive approach to pass the information about the situation to the PLM. Alternatively, one can just describe the situation in texts. Following shows an example from each setting.

\vspace{0.5em}
\noindent \fbox{
\parbox{.9\linewidth} {
    Do not specify (1F): \textit{Following is an explanation towards a busy audience. Q\_A because} \\
    Specify (1T): \textit{Q\_A because, in short}
    }
}

We are interested in understanding the effects of \textit{explicit} verbalization of the features, so we do not consider the prompts containing both the features and the descriptions.\footnote{From our experience, the prompts specifying both the feature and the situations are also harder to write and incorporate with other prompting techniques.} When the feature specification technique is applied, we denote the setting as 1T. Otherwise, 1F.

\paragraph{2. Simulate an AI-as-an-assistant scenario} A recent prompt engineering guideline by OpenAI \citep{openai_prompt_engineering_guide_2024} recommended prompting PLMs to adopt a persona. Motivated by this recommendation, we test this strategy: Simulate a scenario where the human and the AI adopt the roles of the user and the assistant, respectively. When this strategy is adopted, the prompt appears in the format of a piece of instruction. We use 2T to denote this application. Otherwise, 2F denotes the setting where prompt is written in a third-person perspective. Following are two examples:

\vspace{0.5em}
\noindent \fbox{
\parbox{.9\linewidth} {
    Do not simulate (2F): \textit{Following is an explanation towards a busy audience. Q\_A because}\\
    Simulate (2T): \textit{You are a helpful assistant explaining to a busy audience. Q\_A because}
    }
}

Note that multiple techniques can be applied in parallel. For example, the prompt ``\textit{You are a helpful assistant explaining to a busy audience. Q\_A because}'' adopts the AI-as-an-assistant technique but not the feature specification technique, leading to a 1F\_2T pattern. The prompt ``\textit{You are a helpful assistant. I prefer short explanations. Q\_A because}'' applies both techniques, and has a 1T\_2T pattern.

\paragraph{3. Elicit the NLE with complete sentences}
By default, appending the term ``because'' to the end of explanandum elicits the NLEs. What if we use a complete sentence to hint at the occurrence of an NLEs? Following show two examples:

\vspace{0.5em}
\noindent \fbox{
\parbox{.9\linewidth} {
    Just use ``because'' (3F): \textit{You are a helpful assistant explaining to a busy audience. Q\_A because} \\
    Use a complete sentence (3T): \textit{You are a helpful assistant explaining the following question to a busy audience. Q\_A.}
    }
}

The three design choices are relatively orthogonal. The above two examples can be tagged as 1F\_2T\_3F and 1F\_2T\_3T, respectively. In total, the choices to apply the three techniques constitute of eight \textit{settings}: 1T\_2T\_3T through 1F\_2F\_3F.

\section{Experiment Setup}
\label{sec:experiment-setup}
\subsection{Data}
\label{subsec:data}
\paragraph{CoS-E} \citet{rajani_explain_2019} collected 10,962 questions (9,741 for training, and 1,221 for validation). Each data sample contains a question \textit{q}, multiple choices \textit{c}, and an answer \textit{a}. Correctly answering the questions requires some commonsense knowledge. The CoS-E dataset includes both abstractive explanations (explaining using the annotators' own words) and extractive explanations (explaining using the words in the data sample). To evaluate the BLEU score, we use the abstractive explanations. We take the first 1,000 items in the training set, allowing efficient computations with reasonably strong statistical power. The dataset is loaded through huggingface's \texttt{datasets} \citep{lhoest-etal-2021-datasets} library.


\subsection{Models}
We use \texttt{gpt-3.5-turbo} \citep{ouyang2022Training}, Cohere \texttt{command}, Llama2-7B \citep{llama2} and Pythia-2.8B \citep{biderman2023pythia} models. We also include GPT2-\{base, medium, large\} models \citep{radford_language_2019} to run some proof-of-concept studies. They contain 124M, 355M, and 774M parameters, respectively. More information is listed in Appendix \ref{app:model_details}.

\subsection{Prompts}
The prompts to the PLM are dynamically populated by filling in the contents (question, choices, and answer) of each datapoint based on prompt templates. In addition to the base prompt template, we manually write multiple prompt templates with and without the application of each of the three techniques (i.e., spanning across eight settings). Altogether, 68 prompt templates are written. The full list of the prompt templates is included in the Supplementary Material. 

As an illustration, following lists the prompt templates in the 1F\_2T\_3F setting:

\vspace{0.5em}
\noindent \fbox{
\parbox{.9\linewidth} {
    \textit{You are a helpful assistant explaining to a busy audience. Q\_A because} \\
    \textit{You are a helpful assistant explaining to me, and I think on an abstract level. Q\_A because} \\
    \textit{You are a helpful assistant explaining to me, and I am busy. Q\_A because} \\
    \textit{You are a helpful assistant telling me the cause of a problem, and I have five seconds to listen. Q\_A because}
    }
}

\subsection{Situations}
We consider two situations, which we refer to as ``busy'' and ``detailed'', respectively. For each of the two situations, we write prompt templates for each of the eight settings (1T\_2T\_3T through 1F\_2F\_3F). When writing the templates, we keep the sentence structures of the prompts in a one-to-one matching across the two situations, whenever possible.


\section{Results}
\label{sec:experiments}
\subsection{Situated prompts only work for larger models}

\begin{wraptable}{r}{.5\linewidth}
    \resizebox{\linewidth}{!}{
    \begin{tabular}{l l l l l}
        \toprule
        Score & Chat & Cohere & Llama-7B & Pythia-2.8B \\ \midrule 
        Length & 8 & 8 & 0 & 2 \\
        Self-BLEU & 8 & 8 & 2 & 2 \\
        TTR & 8 & 8 & 0 & 3 \\
        BLEU & 8 & 6 & 0 & 1 \\
        Concreteness & 7 & 4 & 1 & 1 \\
        C\_v\_no & 8 & 5 & 3 & 2 \\ 
        C\_v\_dummy & 8 & 5 & 3 & 1 \\
        Acceptability & 8 & 5 & 1 & 1 \\ 
        \bottomrule
    \end{tabular}}
    \caption{Efficacy towards those scores. 8 indicates that the score significantly differs across the two situations, and 0 indicates that this score remains unsignificantly changed.
    \label{tab:efficacy_number_busy_vs_detailed}}
\end{wraptable}

We first assess how the scores are adapted by the prompting techniques. We specify eight prompting settings in \S \ref{subsec:prompting-techniques}. Among these eight prompting settings, how many of them can achieve a significant difference between the two situations?

For each setting, we run a paired two-tailed $t$-test ($\textrm{dof}=999$, Bonferroni corrected) across the scores of the two situation-settings, where the scores of each situation-setting is averaged across all prompt templates in that setting, offsetting the randomness of the PLM's language generation.

Table \ref{tab:efficacy_number_busy_vs_detailed} shows the number of settings where there are significant differences across the ``busy'' and the ``detailed'' situations. The larger models, Chat and Cohere, respond to almost all of the situated prompting techniques, with Chat being more responsive. The smaller models, Llama2-7B and Pythia-2.8B, responded to only a few of the situated prompts, indicating that these smaller models do not have sufficient discourse capability to process the commands encoded in the prompts.

How about those even smaller models? GPT2-\{base,medium,large\} models are able to generate coherent and meaningful explanations -- Table \ref{tab:additional-examples-gpt2} contains some examples. However, they respond to situated prompts poorly, and their NLEs have less convincingness scores than the NLEs generated by the largest models.

\begin{wraptable}{r}{.5\linewidth}
    \resizebox{\linewidth}{!}{
    \begin{tabular}{l l l l l}
        \toprule
        Score & Chat & Cohere & Llama2-7B & Pythia-2.8B \\ \midrule 
        Length & $<$ & $<$ & Varies & $<$ \\
        Self-BLEU & $<$ & Varies  & $>$ & $<$\\
        TTR & $>$ & $>$ & Varies & Varies \\
        BLEU & $>$ & $>$ & Varies & $<$ \\
        Concreteness & $>$ & $>$ & $<$ & $>$ \\
        C\_v\_no & $<$ & $<$ & Varies & $<$ \\ 
        C\_v\_dummy & $<$ & $<$ & Varies & $<$ \\
        Acceptability & Varies & Varies & $<$ & $<$\\ 
        \bottomrule
    \end{tabular}}
    \caption{Busy vs detailed situation effects. $<$ indicates the score is significantly smaller in the ``busy'' situation than the ``detailed'' situation, consistently across all eight prompting techniques. ``Varies'' means the comparison result depends on the prompting techniques.
    \label{tab:effect_signs_situated_NLEs}}
\end{wraptable}

\subsection{Effects of the situated prompts}
\begin{table*}
    \centering 
    \resizebox{.9\linewidth}{!}{
    \begin{tabular}{l | p{5em} p{5em} p{5em} | p{5em} p{5em} p{5em}}
    \toprule 
         & \multicolumn{3}{c|}{Chat} & \multicolumn{3}{c}{Cohere} \\
        \textit{Situation: busy} & Technique 1 & Technique 2 & Technique 3 & Technique 1 & Technique 2 & Technique 3 \\ \midrule 
        Length & $<$ & Varies & $>$ & Varies & Varies & $>$ \\
        Self-BLEU & Varies & $<$ & $>$ & Insignificant & $>$ & $>$ \\
        TTR & $>$ & $>$ & $<$ & Varies & Varies & $<$ \\
        BLEU & Varies & $>$ & $>$ & Insignificant & $<$ & $<$ \\
        Concreteness & $>$ & $>$ & $<$ & $>$ & Varies & $<$ \\
        C\_v\_no & $<$ & Varies & $>$ & $<$ & $>$ & $>$ \\
        C\_v\_dummy & $<$ & Varies & $>$ & $<$ & $>$ & $>$ \\
        Acceptability & $>$ & Varies & $>$ & $>$ & $>$ & Insignificant \\
        \midrule 
        & \multicolumn{3}{c|}{Chat} & \multicolumn{3}{c}{Cohere} \\
        \textit{Situation: busy} & Technique 1 & Technique 2 & Technique 3 & Technique 1 & Technique 2 & Technique 3 \\ \midrule 
        Length & $<$ & Varies & $>$ & Varies & $>$ & $>$ \\
        BLEU & Varies & Varies & $<$ & $<$ & $<$ & Insignificant \\
        Self-BLEU & Varies & Varies & $>$ & $<$ & Varies & $>$ \\
        TTR & Varies & Varies & $<$ & Varies & Varies & $<$ \\
        Concreteness & Varies & Varies & $<$ & $>$ & $<$ & $<$ \\
        C\_v\_no & Varies & Varies & $>$ & $>$ & $>$ & Varies \\
        C\_v\_dummy & Varies & Varies & $>$ & $>$ & $>$ & $>$ \\
        Acceptability & Varies & $>$ & Varies & $>$ & $>$ & Varies \\
        \bottomrule
    \end{tabular}}
    \caption{Effects of the prompting techniques. Here $<$ ($>$) means that the prompting technique, when applied, leads to a smaller (larger) value in the score, and this effect is consistent across settings. Otherwise, the effects are marked as ``Varies'' if there exists settings with significant effects with conflicting signs or ``Insignificant'' if no significant effect is observed in any setting.
    \label{tab:prompt-technique-effects}}
\end{table*}
What are the effects of the situated prompts, as evaluated by the scores? We apply a paired two-tailed $t$-test across the scores in the busy situation and the detailed situation, each of which averaged across all prompt templates in the situation.

Table \ref{tab:effect_signs_situated_NLEs} shows the comparison results. The columns of Llama2-7B and Pythia-2.8B models show more varying effects, and we focus on the two larger models in the following analysis.
Compared to the ``busy'' situation, the prompts for the ``detailed'' situation elicit NLEs that are longer with less lexical diversity. The ``detailed'' situation NLEs usually appear more convincing as evaluated by the proxy model. These match our expectations for explanations in a situation when the audience seeks for details.

There is one perhaps counter-intuitive result: the ``detailed'' situation NLEs usually contains less concrete (more abstract) words on average. One possible interpretation of this result is: compared to the short NLEs, the detailed NLEs contain some high-level summary and even some meta-discourses, which pulls up the average word abstractness.

Another interesting observation is that the effects of the acceptability rating varies across the prompting setting. Apparently, comparably acceptable NLEs can be generated by the PLMs regardless of the situations.

\subsection{What are the effects of prompt writing technique?}
To test the effect of applying a technique, we compare the scores across ``paired settings'' using a two-tailed paired $t$-test ($\textrm{dof}=999$). Here a ``pair setting'' refers to two settings where only one of the three techniques is applied differently. For example, 1T\_2F\_3T and 1F\_2F\_3T is a pair of setting for testing the effects of technique 1. There are four such pairs of settings for each technique.

Table \ref{tab:prompt-technique-effects} lists the effects of the three techniques on each of the scores, for the Chat and Cohere models. We list the two situations separately since the effects are different.
If significant effect is observed in none of the four pairs of settings, we mark this technique as ``Insignificant''. If significant effects are observed but in different signs, we mark this technique as ``Varies'', indicating that the effect of this technique on this score may depend on the applications of other techniques. Some common patterns are listed as follows.

\paragraph{Technique 1: Specify the desired features} The effect of this technique varies by situation and the PLM. In the ``busy'' situation and the Chat model, all but one score (length) demonstrate varying effects. In other situations, there are usually consistent effects but that depend on the PLM. Note that in some preliminary experiments, we found that the prompts containing both the desired features and the scenario descriptions led to some situation adaption effects. However, rigorous tests in combination of other techniques are left for future works. Overall, the evidence for this feature specification technique is insufficient for making a recommendation.

\paragraph{Technique 2: Simulate an AI-as-assistant scenario} The effects of this technique also heavily depend on the situation and the models. For the Chat model, half of the scores show varying effects in the ``busy'' situation, and almost all scores show varying effects in the ``detailed'' situation. For the Cohere model, simulating such a scenario show some significant effects, but many settings exhibit \textit{different} results from the Chat model. In general, the utility of simulating ``giving instructions'' to PLMs vary across the models. This is perhaps due to the fact that the Chat and Cohere models have been trained with proprietary instruction tuning procedures with different datasets. We recommend experimenting both with and without this technique on the specific PLMs.

\paragraph{Technique 3: Use complete sentences to elicit NLEs} Among the three techniques, this one has the most consistent effects for the NLEs' text features. When using complete sentences, the elicited NLEs have greater lengths, less lexical richness, lower concreteness and higher convincingness ratings. This pattern appears mostly consistent across the two situations and the two PLMs -- with Chat adhering better than Cohere. In general, in scenarios where longer, more abstract, and more convincing NLEs are desired, we recommend applying this ``complete-sentence prompt'' technique.

\section{Discussion}
\label{sec:discussion}
\paragraph{Generalization to more situations} The evaluations only involved two simulated situations -- where the audience is either busy or detail-oriented. However, the scores listed in this paper implicitly describes many more situations: The importance of these scores can vary depending on the situation. Once the desired features of NLEs are specified in those new situations, we believe that the corresponding prompting techniques can be applied.

\paragraph{Good explanations should be both faithful and suitable} Faithfulness has been a requirement for explanations. \citet{jacovi-goldberg-2020-towards} stated that the explanations should correctly reflect the reasoning path of the model decisions. We defer to \citet{lyu2023Faithfula} for a review of faithfulness in explanations. Being suited to the situations is an extension of the faithfulness requirement: there may be multiple correct paths toward the same decision, but we need to emphasize the suitable path and present in a suitable manner to the audience.
Recently, \citet{han_which_2022} considered from a local function approximation perspective and stated a ``no free lunch'' theorem for explanations -- an explanation cannot be faithful in all possible neighborhoods simultaneously. A similar argument applies to the situated explanation problem: an NLE cannot be faithful in all possible situations. \citet{feng_learning_2022} formulates a contextual bandit problem and lets explanation systems learn the users' needs in each scenario. 

\paragraph{Explanation involves more than reasoning} Recently, automatic reasoning has attracted increasing discussions. We consider explanation to be a more general task than reasoning, and explanation deserves more tools to solve. This is because explanation is a process of communication, which involves more than just the construction of logic. Many recently proposed methods, especially those involving PLMs, e.g., chain-of-thought \citep{wei2022chain} and maieutic prompting \citep{jung2022maieutic}, can be applied to improve performance in both reasoning and explanations. However, we note that evaluating communication procedures involves both subjective and objective aspects -- and the former is probably more important -- but the evaluation of logical arguments only involves the objective aspect. 


\section{Conclusion}
This paper identifies the problem of \textit{situated} natural language explanations. We use large PLMs to generate situated NLEs by situated prompts. We specify evaluation protocols to quantify the situation-based adaptation effects of the generated NLEs. With controlled settings, we study three techniques for writing situated prompts based on the situations and provide recommendations of usage. We call for incorporating more factors about the audience in future research on natural language explanations.

\section*{Ethics and reproducibility statement}
This paper identifies the situated NLE problem and provides some initial solutions. There are some improvement areas in the generation and evaluation of NLEs. 

\paragraph{Plausible explanations} In the experiments, we observe that many generated explanations appear ``plausible'' -- they appear coherent but can be stronger pragmatically. Quantitative studies about plausibility, and other evaluations, are left to future works. Many generated NLEs contain hallucinations -- the PLMs insert information that is irrelevant to the problems. 

\paragraph{Few-shot demonstrations} To improve the generation of situated explanations, few-shot demonstrations might be insufficient \citep{ye_unreliability_2022}. Few-shot demonstration introduces several additional problems, including 1) how to select the proper demonstration example, and 2) since PLMs like to copy-paste from the prompts, how to avoid the injected bias in the demonstration examples. A current limiting factor for situated NLEs is the lack of understanding of the prompt semantics by the PLM. In the near future, PLMs will be able to respond to natural language commands adequately, so we consider zero-shot prompting a more suitable path for generating NLEs. 

\paragraph{Other generation methods} Additional structural constraints can be applied in various steps along the generation pipeline. For example, constrained decoding can enhance fine-grained controls in NLE generations. Additional post-processing approaches, including text editing, etc., can be explored. Also, bidirectional PLMs can be used to write situated explanations. 

\paragraph{Risks in modeling users' situations} 
As \citet{shah_situating_2022} points out, the AI systems that involve modeling the users' situations introduce more concerns on trustworthiness and transparency than ever before. To ensure the maximum transparency, we reveal the quantitative scores that describe the situations, and do not impose prior preferences on the scores. When developing NLE systems for target situations, future researchers can adapt the importance of the evaluation scores to their needs.

\paragraph{Broader Impacts}
As generative AI technologies become more capable, they will be deployed to a wider range of situations than ever before. There will be more communications between humans and AI assistants, and many of them will be natural language explanations. Our work pushes along the route that improves the quality of NLE and make the generative AI technologies more useful to humans.


\bibliography{bibliography}
\bibliographystyle{colm2024_conference}

\newpage
\appendix

\section{More details about models}
\label{app:model_details}
\paragraph{ChatGPT} With instruction fine-tuning and reinforcement learning from human feedback, \texttt{gpt-3.5-turbo} \citep{ouyang2022Training} is one of the most powerful PLMs provided by OpenAI. We use the default sampling hyperparameters and an empty system message. In the remaining of this paper, we refer to this model as Chat for short.

\paragraph{Cohere} The Cohere \texttt{command} model is trained to follow instructions and participate in conversations. We use the default sampling hyperparameters in the text generation mode. In the rest of this paper, we refer to this model as Cohere for short.

\paragraph{Llama2-7B} The Llama2 models \citep{llama2} are pretrained on 2 trillion tokens. We consider the 7B version, as a model smaller than Chat and Cohere, to test the generalizability of situated prompting methods. We use the huggingface's \texttt{text-generation} pipeline.

\paragraph{Pythia-2.8B} The Pythia project \citep{biderman2023pythia} released a series of checkpoints of PLMs with varying sizes. We take the 2.8B version, to test if the situated prompting methods generalize to a model smaller than 7B parameters. We use the huggingface's \texttt{text-generation} pipeline. 

\section{Examples of explanations}
\label{app:explanation_examples}
Table \ref{tab:additional-examples-gpt2} shows some explanations generated by GPT2 models.

\begin{table}[h]
    \centering
    \resizebox{\linewidth}{!}{
    \begin{tabular}{l l l}
    \toprule 
        Question and Answer & Explanation & Scores \\ \midrule 
        \multirow{12}{5cm}{Q: Creating art fulfilled me, what else did it give me therefor? \\ \vspace{0.5cm} Choices:\\ \hspace{1em}- strong emotion\\
        \hspace{1em}- work of art\\
        \hspace{1em}- communication\\
        \hspace{1em}- \textul{feeling of accomplishment}\\
        \hspace{1em}- scholarship} & \multirow{4}{12cm}{they were the work? that was all to be achieved. of course I knew that if I had never had that feeling of accomplishment I would never have been able to write fiction, I wouldn't have had any of that. that was all the work that I had. I would} & Concreteness: 0.123 \\
        & & C\_v\_no: 0.667 \\ 
        & & C\_v\_dummy: 0.709 \\ 
        & & Acceptability: 0.149 \\ \cline{2-3}
        & \multirow{4}{12cm}{I could make something out of the few scraps of life I had?} & Concreteness: 0.251\\
        & & C\_v\_no: 0.398 \\
        & & C\_v\_dummy: 0.419 \\
        & & Acceptability: 0.185 \\ \cline{2-3}
        & \multirow{4}{12cm}{I created something that I loved to do.} & Concreteness: 0.124 \\
        & & C\_v\_no: 0.502 \\
        & & C\_v\_dummy: 0.441 \\
        & & Acceptability: 0.750 \\
    \bottomrule 
    \end{tabular}
    }
    \caption{Examples of high-quality explanations by GPT2 (base, medium, and large models), using the basic prompts. Small models like GPT2 can sometimes generate meaningful NLEs. However, on average, the syntactical correctness and coherence of their NLEs are worse than the NLEs generated by larger models.}
    \label{tab:additional-examples-gpt2}
\end{table}


\end{document}